\documentclass[11pt,letterpaper]{article}
\usepackage{emnlp2017}
\usepackage{times}
\usepackage{latexsym}

\usepackage{url}
\usepackage{mathrsfs}
\usepackage{amsmath}
\usepackage{amssymb}
\usepackage{courier}
\usepackage{graphicx}
\usepackage{multirow}
\usepackage{color}
\usepackage{indentfirst}
\usepackage{floatrow}
\usepackage{scrextend}
\newfloatcommand{capbtabbox}{table}[][\FBwidth]

\emnlpfinalcopy

\def\emnlppaperid{1333}

\title{Incorporating Relation Paths in Neural Relation Extraction}

\author{Wenyuan Zeng$^1$, Yankai Lin$^2$, Zhiyuan Liu$^{2}\thanks{$^*$Corresponding author: Z. Liu (liuzy@tsinghua.edu.cn).}$, Maosong Sun$^2$\\
  $^1$Department of Physics, Tsinghua University, Beijing, China\\
  $^2$State Key Laboratory of Intelligent Technology and Systems\\
  Tsinghua National Laboratory for Information Science and Technology\\
  Department of Computer Science and Technology, Tsinghua University, Beijing, China}

\date{}

\begin{document}

\maketitle
\begin{abstract}
Distantly supervised relation extraction has been widely used to find novel relational facts from plain text. To predict the relation between a pair of two target entities, existing methods solely rely on those direct sentences containing both entities. In fact, there are also many sentences containing only one of the target entities, which also provide rich useful information but not yet employed by relation extraction. To address this issue, we build inference chains between two target entities via intermediate entities, and propose a path-based neural relation extraction model to encode the relational semantics from both direct sentences and inference chains. Experimental results on real-world datasets show that, our model can make full use of those sentences containing only one target entity, and achieves significant and consistent improvements on relation extraction as compared with strong baselines. The source code of this paper
can be obtained from \url{https://github.com/thunlp/PathNRE}.

\end{abstract}

\section{Introduction}
Knowledge Bases (KBs) provide effective structured information for real world facts  and  have  been  used as crucial  resources  for  several  natural language processing (NLP) applications such as  Web search and question answering. Typical KBs such as Freebase \cite{bollacker2008freebase}, DBpedia \cite{auer2007dbpedia} and YAGO \cite{suchanek2007yago}  usually describe knowledge as multi-relational data and represent them as triple facts. As the real-world facts are infinite and increasing every day, existing KBs are still far from complete. Recently, petabytes of natural-language text containing thousands of different structure  types  are  readily  available, which is an important resource for  automatically finding unknown relational facts. Hence, relation  extraction  (RE),  defined  as  the  task  of extracting  structured  information   from plain text, has attracted much interest.

Most existing supervised RE systems usually suffer from the issue that lacks sufficient labelled relation-specific training  data. Manual annotation is very time consuming and labor intensive. One promising  approach  to address this limitation is distant supervision. \cite{mintz2009distant} generates training data automatically by aligning a KB with plain text.  They assume that if two target entities have a relation in KB, then all sentences that contain these two entities will express this relation and can be regarded as a positive training instance. Since neural models have been verified to be effective for classifying relations from plain text \cite{socher2012semantic,zeng2014relation,dos2015classifying}, \cite{zeng2015distant, lin2016relation} incorporate neural networks method with distant supervision relation extraction. Further, \cite{ye2016jointly} considers finer-grained information, and achieves the state-of-the-art performance.

Although existing RE systems have achieved promising results with the help of distant supervision and neural models, they still suffer from a major drawback: the models only learn from those sentences contain both two target entities. However, those sentences containing only one of the entities could also provide useful information and help build inference chains. For example, if we know that ``$h$ is the father of $e$" and ``$e$ is the father of $t$", we can infer that $h$ is the grandfather of $t$. 

 \begin{figure*}[!htp]
\centering
\includegraphics[width=0.75\textwidth]{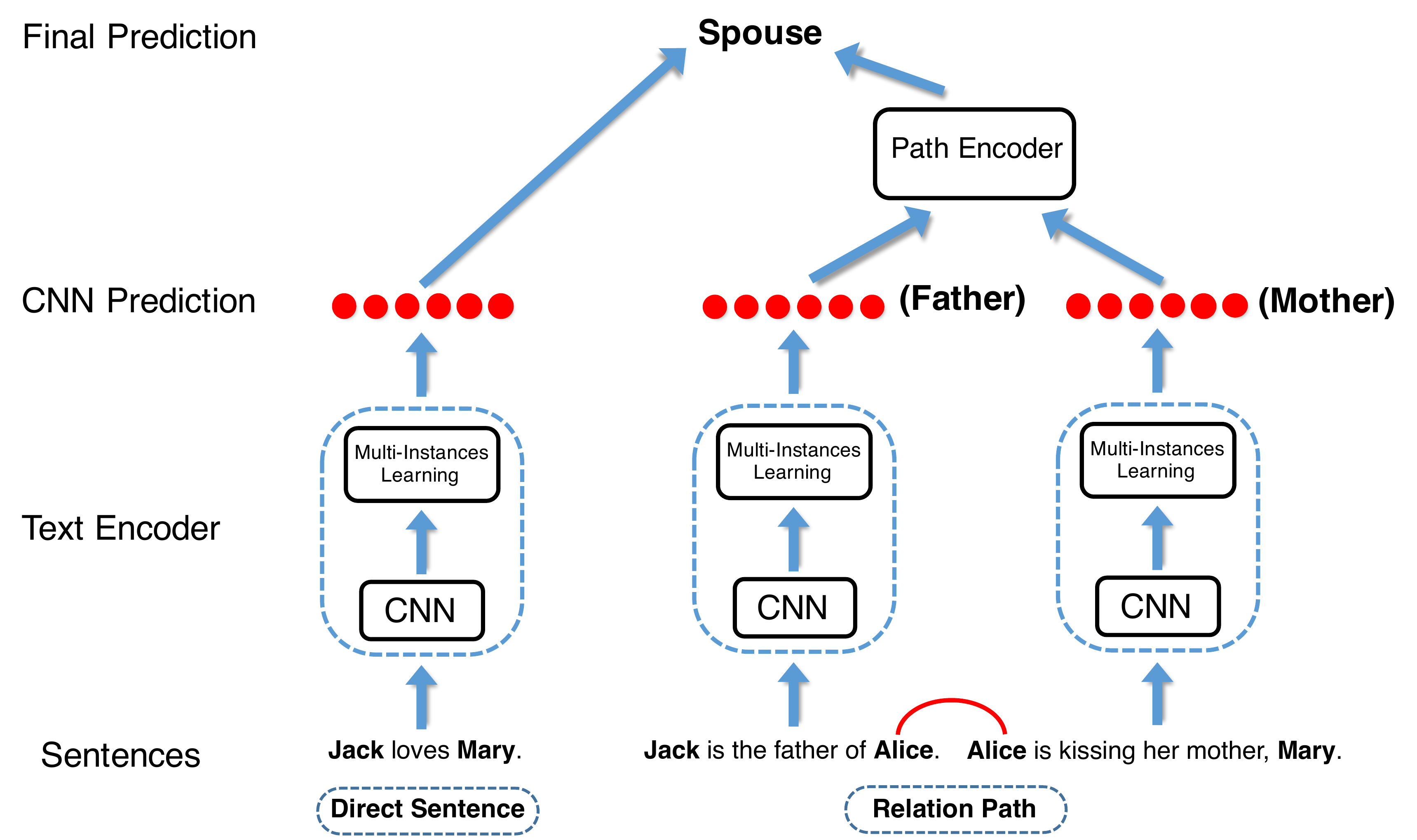}
\caption{The architecture of our neural relation extraction model with relation paths.}
\label{fig:model}
\end{figure*}

In this work, as illustrated in Fig. \ref{fig:model}, we introduce a path-based neural relation extraction model with relation paths. First, we employ convolutional neural networks (CNN) to embed the semantics of sentences. Afterward, we build a relation path encoder, which measures the probability of relations given an inference chain in the text. Finally,  we combine information from direct sentences and relation paths to predict the relation.  

We evaluate our model on a real-world dataset for relation extraction. The experimental results show that our model achieves significant and consistent improvements as compared with baselines. Besides, with the help of those sentences containing one of the target entities, our model is more robust and performs well even when the number of noisy instances increases. To the best of our knowledge, this is the first effort to consider the information of relation path in plain text for neural relation extraction. 


\section{Related Work}


\subsection{Distant Supervision}
Distant supervision for RE is originally proposed in \cite{craven1999constructing}. They   focus  on  extracting  binary  relations  between  proteins using a  protein KB as the source of distant supervision.  Afterward, \cite{mintz2009distant} aligns plain text with Freebase, by using distant supervision . However, most of these methods heuristically transform distant supervision to traditional supervised learning, by regarding it as a single-instance single-label problem, while in reality, one instance could correspond with multiple labels in different scenarios and vice versa. To alleviate the issue, \cite{riedel2010modeling}  regards each sentence as a training instance and  allows  multiple instances to share  the same label but disallows more than one label. 
Further, \cite{hoffmann2011knowledge,surdeanu2012multi}  adopt multi-instance multi-label learning  in relation extraction. 
The main drawback of these methods is that they obtain most features directly from NLP tools with inevitable errors, and these errors will propagate to the relation extraction system and limit the performance.

\subsection{Neural Relation Extraction}
Recently, deep learning \cite{bengio2009learning} has  been successfully applied in various areas, including computer vision, speech recognition and so on. Meanwhile, its effectiveness has also been verified in  many NLP tasks such as sentiment analysis \cite{dos2014deep}, parsing \cite{socher2013parsing}, summarization \cite{rush2015neural} and machine translation \cite{sutskever2014sequence}. With the advances of deep learning, there are growing works that design neural networks for relation extraction. \cite{socher2012semantic} uses a recursive neural network in relation extraction, and \cite{xu2015classifying, miwa2016end} further use LSTM. \cite{zeng2014relation,dos2015classifying} adopt CNN in this task, and \cite{zeng2015distant, lin2016relation} combine attention-based multi-instance learning which shows promising results. However, these above models merely learn from those sentences which directly contain both two target entities. The important information of those relation paths hidden in the text is ignored. In this paper, we propose a novel path-based neural RE model to address this issue. Besides, although we choose CNN to test the effectiveness of our model, other neural models could also be easily adapted to our architecture.



\subsection{Relation Path Modeling}
Relation paths have been taken into consideration on large-scale KBs for relation inference. Path Ranking algorithm (PRA) \cite{lao2010relational} has been adopted for expert finding \cite{lao2010relational}, information retrieval \cite{lao2012reading}, and further for relation classification based on KB structure \cite{lao2011random,gardner2013improving}. \cite{neelakantan2015compositional,lin2015modeling,das2016chains,wu2016knowledge} use recurrent neural networks (RNN) to represent relation paths based on all involved relations in KBs.\cite{guu2015traversing} proposes an embedding-based compositional training method to connect the triple knowledge for KB completion. Different from the above work of modeling relation paths in KBs, our model aims to utilize relation paths in text corpus, and help to extract knowledge directly from plain text.

\section{Our Method}
Given a pair of target entities, a set of corresponding direct sentences $S = \{s_1,s_2,\cdots,s_n\}$ which contains this entity pair, and a set of relation paths $P = \{p_1, p_2, \cdots, p_m\}$,   our model aims to measure the confidence of each relation for this entity pair. In this section, we will introduce our model in three parts: (1) \textbf{Text Encoder.} Given the sentence with two corresponding target entities,  we use a CNN  to embed the sentence into a semantic space, and measure the probability of each relation given this sentence. (2) \textbf{Relation Path Encoder.} Given a relation path between the target entities, we measure the probability of each relation $r$, conditioned on the relation path. (3) \textbf{Joint Model.} We integrate the information from both direct sentences and relation paths, then predict the confidence of each relation.

\subsection{Text Encoder}

\begin{figure}[!htp]
\centering
\includegraphics[width=1.0\columnwidth]{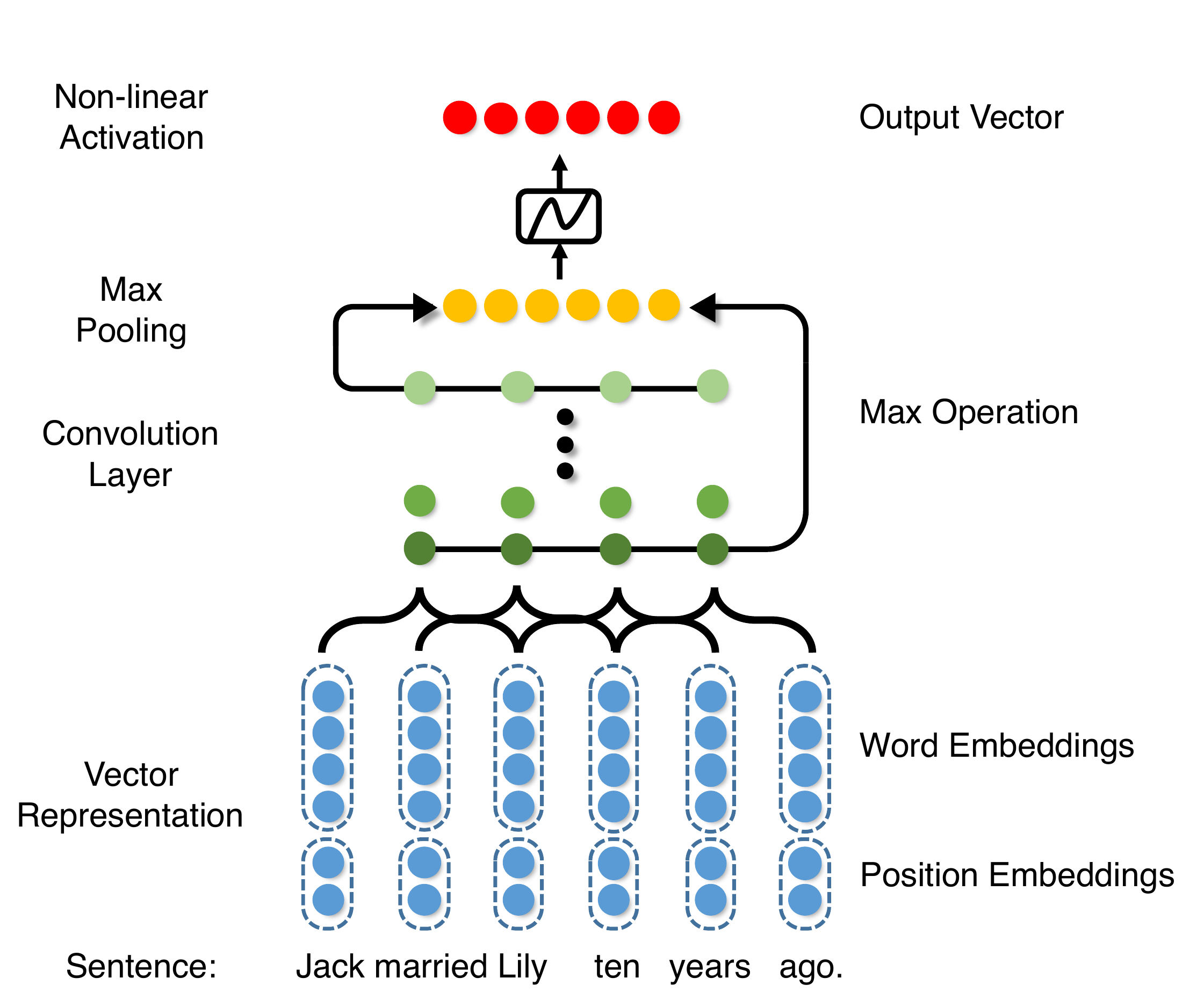}
\caption{The architecture of CNN used for text encoder.}
\label{fig:cnn-model}
\end{figure}

 As shown in Fig. \ref{fig:cnn-model}, we use a  CNN to extract information from text. 
Given a set of sentences of an entity pair, we first transform each sentence $s$ into its distributed representation 
$\mathbf{s}$, and then predict relation using the most representative sentence via a multi-instance learning mechanism.
 
 \subsubsection{Input Vector}
 First, we transform the words $\{w_1,w_2,\cdots,w_l\}$ in sentence $s$ into vectors of dimension $d$. For each word $w_i$, we use word embedding to encode its syntactic and semantic meanings, and use position embedding to encode its position information. We then concatenate both word embedding and position embedding to form the input vector of $w_i$ for CNN. (See Figure \ref{fig:cnn-model}.)

 \subsubsection{Convolution and Max-pooling Layers}
 When processing a sentence, it is a great challenge that important information could probably appear in all parts of that sentence. In addition, the length $l$ of a sentence could also vary a lot. Therefore, we apply CNN to encode all local features regardless  sentence length. We first apply a convolution layer to extract all possible local features, and then select the most important one via max-pooling layer. 
 
 To extract local features, the convolution layer first concatenates a sequence of word embeddings within a sliding window to be vector $\mathbf{q}_i $ of dimension $k \times d$:
\begin{equation}
\mathbf{q}_i = \mathbf{w}_{[i-k+1:i]} (1\leq i \leq l+k-1),
\end{equation}
where $k$ is the size of the window, and we also set all out-of-index words to be zero vectors. It then multiplies $\mathbf{q}_i$ by a convolution matrix $\mathbf{W}\in\mathbb{R}^{d_c \times (k \times d)}$, where $d_c$ is the dimension of sentence embeddings. Hence, the output of convolution layer could be expressed as $\mathbf{h} = \{\mathbf{h}_1,\mathbf{h}_2,\cdots,\mathbf{h}_{l+k-1}\}$:
  
\begin{equation}
\mathbf{h}_i = \mathbf{W} \mathbf{q}_i+\mathbf{b},
\end{equation}
where $\mathbf{b}$ is a bias vector. Finally, the max-pooling layer takes a max operation, followed by a hyperbolic tangent activation, over the sequence of $\mathbf{h}_i$ to select the most important information, namely,
  
\begin{equation}
[\mathbf{s}]_j = \mathbf{tanh}(\max\limits_{i}[\mathbf{h}_i]_j).
\end{equation}

 \subsubsection{Multi-Instance Learning}
 Next, we apply a softmax classifier upon the sentence representation $\mathbf{s}$ to  predict the corresponding relation. We  define the condition probability of relation $r$ as follows,
  \begin{equation}
 p(r|\theta,\mathbf{s}) = \frac{\exp(e_r)}{\sum^{n_r}_{i = 1}{\exp(e_i)}},
 \end{equation}
 where $e_i$, a component of $\mathbf{e}$, measures how well this sentence matches relation $r_i$, and $n_r$ is the number of relations. More specifically, $\mathbf{e}$ could be calculated from:
\begin{equation}
\mathbf{e} = \mathbf{U}\mathbf{s}+\mathbf{v},
\end{equation}
where $\mathbf{U} \in \mathbb{R}^{n_r \times d_c}$ is the coefficient matrix of relations and $\mathbf{v} \in \mathbb{R}^{n_r}$ is a bias vector.
 

We use multi-instance learning to alleviate the wrong-labeling issue in distant supervision, by choosing one sentence in the set of all direct sentences  $S = \{s_1,s_2,\cdots,s_m\}$ which corresponds to the entity pair $(h, t)$. Similar to \cite{zeng2015distant}, we define the score function of this entity pair and its corresponding relation $r$ as a max-one setting:
\begin{equation}
E(h,r,t|S) = \max \limits_i p(r|\theta,\mathbf{s}_i).
\end{equation}
where $E$ reflects the direct information we derive from sentences. We can also set a random setting as a baseline:
\begin{equation}
E(h,r,t|S) = p(r|\theta,\mathbf{s}_i),
\end{equation}
where $s_i$ is randomly selected  from $S$.

 \subsection{Relation Path Encoder}
We use Relation Path Encoder to embed the inference information of relation paths.
 Relation Path Encoder measures the probability of each relation $r$ given a relation path in the text. This will utilize the inference chain structure to help make predictions. More specifically, we define a path $p_1$ between $(h,t)$ as $\{(h,e),(e,t)\}$, and the corresponding relations are $r_{A}$, $r_{B}$. Each of $(h,e)$ and $(e,t)$ corresponds to at least one sentence in the text. Our model calculates the probability of relation $r$ conditioned on $p_1$ as follows,
\begin{equation}
p(r|r_{A},r_{B}) = \frac{\exp(o_r)}{\sum^{n_r}_{i = 1}\exp(o_i)},
\end{equation}
where $o_i$  measures how well relation $r$ matches with the relation path $(r_{A},r_{B})$. Inspired by the work on relation path  representation learning \cite{lin2015modeling}, our model first transforms relation $r$ to its distributed representation, i.e. vector $\mathbf{r} \in  \mathbb{R}^{d_R}$, and builds the path embeddings by composition of relation embeddings. Then, the similarity $o_i$ is calculated as follows:
\begin{equation}
o_i = -\| \mathbf{r}_i - ( \mathbf{r}_{A}+ \mathbf{r}_{B})\|_{L_1}.
\end{equation}

  Therefore, if $\mathbf{r}_i$ gets more similar to $( \mathbf{r}_{A}+ \mathbf{r}_{B})$, the conditioned predicting probability of $r_i$ will become larger. Here, we make an implicit assumption that if $r_i$ is semantically similar to relation path $p_i:h\xrightarrow{r_A}e\xrightarrow{r_B}t$, the embedding $\mathbf{r}_i$ will be closer to the relation path embedding $( \mathbf{r}_{A}+ \mathbf{r}_{B})$.  Finally, for this relation path $p_i:h\xrightarrow{r_A}e\xrightarrow{r_B}t$, we define an relation-path score function,
\begin{equation}
G(h,r,t|p_i) = E(h,r_A,e)E(e,r_B,t) p(r|r_A,r_B),
\end{equation}
where $E(h,r_A,e)$ and $E(e,r_B,t)$ measure the probabilities of relational facts $(h,r_A,e)$ and $(e,r_B,t)$ from text, and $p(r|r_A,r_B)$ measures the probability of relation $r$ given relation path $(r_A, r_B)$.

In reality, there are usually multiple relation paths between two entities. Hence, we define the inferring correlation between relation $r$ and several sentence paths $P$ as,
\begin{equation}
G(h,r,t|P) = \max \limits_{i}G(h,r,t|p_i),
\end{equation}
where we use max operation to filter out those noisy paths and select the most representative path. 

\subsection{Joint Model}
Given any entity pair $(h,t)$, those sentences  $S$ directly mentioning them and relation paths $P$ between them, we define the global score function with respect to a candidate relation $r$  as,
\begin{equation}
\label{eq:total}
L(h,r,t) = E(h,r,t|S) + \alpha G(h,r,t|P),
\end{equation}
where $E(h,r,t|S)$ models the correlation between $r$ and $(h,t)$ calculated from direct sentences, $G(h,r,t|P)$ models the inferring correlation between relation $r$ and several sentence paths $P$.
$\alpha$ equals to $(1-E(h,r,t|S))$ times a constant $\beta$. This term serves to depict the relative weight between direct sentences and relation paths, since we don't need to pay much attention on extra information when CNN has already given a confident prediction, namely $E(h,r,t|S)$ is large.

One of the advantages of this joint model is to alleviate the issue of error propagation. The uncertainty of information from Text Encoder and Relation Path encoder is characterized by its confidence, and could be integrated and corrected in this joint model step. Furthermore, since we treat relation paths in a probabilistic way, our model could fully utilize all relation paths, i.e. those always hold and those likely to hold.

\subsection{Optimization and Implementation Details}
The overall objective function is defined as:
\begin{equation}
J(\theta) = \sum \limits_{(h,r,t)} \log(L(h,r,t)),
\end{equation}
where the summing runs over the log loss of all entity pairs in text and $\theta$ represents the model parameters. To solve this optimization problem, we use mini-batch stochastic gradient descent (SGD) to maximize our objective function. We initialize $\mathbf{W}_E$ with the results from Skip-gram model, and initialize other parameters randomly. We also adopt dropout \cite{srivastava2014dropout} upon the output layer of CNN.

We implement our model using C++. We train our model on Intel(R) Xeon(R) CPU E5-2620, and the training roughly takes half a day. The word embedding and other parameters are updated via back-propagation simultaneously, while the relation path structure is extracted before training and stored afterward.

\section{Dataset}
We build a novel dataset for evaluating relation extraction task. We first describe the most commonly used previous dataset and then explain the reason and how we construct the new dataset.

\subsection{Previous Datasets \& Reasons for New Dataset}
A commonly used benchmark dataset for this task was developed by \cite{riedel2010modeling}. This dataset was built by aligning Freebase (Dec. 2009 Snapshot) with New York Times corpus (NYT). There are 53 possible relationships between two entities, including a special relation type NA, meaning that there is no relation between head and tail entities. For each relational fact in a filtered Freebase dataset, a sentence from NYT would be regarded as a mention of this relation if both the head and tail entity appear in that sentence. 

While this previous dataset has been frequently used for evaluating relation extraction systems, we observe some limitations of it. First, the relational facts are extracted from a 2009 snapshot of Freebase. Therefore, this dataset is too old to contain many updated facts. This will underestimate the performance of a relation extraction system, since some real-world facts are missing from the dataset and labeled as NA. Second, the relational facts in this dataset are scattered, i.e. there are not sufficient relation paths in this dataset, while relational facts in real-world always have connections with each other. Third, Freebase will no longer update after 2016.These limitations mean that this dataset is somewhat improper for evaluating RE systems.

Although other relation extraction datasets exist, e.g. ACE\footnote{\url{https://catalog.ldc.upenn.edu/LDC2006T06}} and \cite{hendrickx2009semeval}, they are too small to train an effective neural relation extraction model. Moreover, each relational fact in \cite{hendrickx2009semeval} only corresponds with one sentence, which prevents it from evaluating multi-instance relation extraction systems. Hence, we constructed a novel relation extraction dataset to address these issues, and will make it available to the community.

\subsection{Dataset Construction}

\begin{table}[htbp]
\centering
\tiny
\captionsetup[table]{position=below}
\begin{tabular}{|c|c|c|c|c|}
\hline
Datasets & Sets & \# sentences & \# entity pairs & \# facts \\
\hline
\multirow{ 3}{*}{Riedel et.al.} & Train 	&522,611	&281,270	&18,252	 \\
	&Valid &-	&-	&-	 \\
	&Test &172,448	&96,678	&1,950	\\
\hline
\multirow{ 3}{*}{Ours} & Train 	&647,827	&266,118	&50,031  \\
	&Valid &234,350	&121,160	&5,609	\\
	&Test &235,609	&121,837	&5,756	\\
\hline
\end{tabular}
\caption{Statistics of datasets.}
\label{table:data}
\end{table}


Our dataset contains more updated facts and richer structures of relations, e.g. more relations / relation paths, as compared to existing similar datasets. The dataset is expected to be more similar to real-world cases, and thus be more appropriate for evaluating RE systems' performances. 

We build the dataset by aligning Wikidata\footnote{\url{https://www.wikidata.org/}} relations with the New York Times Corpus (NYT). 
Wikidata is a large, growing knowledge base, which contains more than 80 million triple facts and 20 million entities. Different from Freebase, Wikidata is still in maintenance and could be easily accessed by APIs. We first pick those entities simultaneously appeared in both Wikidata and Freebase, and relational facts associated with them. Then, we filtered out a subset $S$, reserving those facts associating with the 99 highest frequency relations. This results in $4,574,665$ triples with $1,045,385$ entities and $99$ relations. 

Next, we align those facts with NYT corpus, following the assumption of distant supervision. For each pair of entities appearing in our $S$, we traverse the corpus and pick those sentences where both entities appear. These sentences will be regarded as mentions of this fact, and labeled by this relation type. To simulate noise in the real world, we also add sentences corresponding to ``No Relation" entity pairs into our dataset. To get those ``No Relation'' instances, we first create a fake knowledge base $S^-$ by randomly replacing the head or tail entities in triples, i.e., $S^{-} = \{(h', r, t)\} \cup  \{(h, r, t')\}$ and then align them with NYT corpus. Finally, we randomly split all those selected sentences into training, validation and testing set, assuring that a relational fact could be only mentioned by sentences in one set. The statistics of our dataset and \cite{riedel2010modeling} are listed in Table \ref{table:data}.


\section{Experiments}

Following the previous work \cite{mintz2009distant}, we evaluate our model by extracting relational facts from the sentences in test set, and compare them with those in Wikidata. We report Precision/Recall curves, Precision@N (P@N) and F1 scores for comparison in our experiments. 

\subsection{Initialization and Parameter Settings}

In this paper, we use the word2vec tool \footnote{\url{https://code.google.com/p/word2vec/}} to pre-train word embeddings on NYT corpus. We keep the words which appear more than 100 times in the corpus as vocabulary. 
We tune our model on the validation set, using grid search to determine the optimal parameters, which are shown in boldface. We select learning rate for SGD $\lambda \in \{0.1,\textbf{0.01},0.001\}$, the sentence embedding size $d_c \in \{50,60,\cdots,\textbf{230},\cdots,300\}$, the window size $k \in \{1,2,\textbf{3},4,5\}$, and the mini-batch size $B \in \{40,\textbf{160},640\}$. Besides, we select the relation embeddings size $d_R \in \{5,10,\cdots,\textbf{40},\cdots,60\}$, and the weight for information from relation paths $\beta \in \{0.01,0.1,0.2,\textbf{0.5},1,\cdots,5\}$. For other parameters which have little effect on the system performance, we follow the settings used in \cite{zeng2015distant}: word embedding size $d_w$ is 50, position embedding size $d_p$ is 5 and dropout rate $p$ is 0.5. For training, the iteration number over all training data is 25. 

\subsection{Effectiveness of Incorporating Relation Paths}
\begin{table*}[htp]
\centering
\small
\begin{tabular}{c|cccc|cccc|cccc}
\hline
Test Settings (Noise) & \multicolumn{4}{c|}{75$\%$}&\multicolumn{4}{c|}{85$\%$}&\multicolumn{4}{c}{95$\%$}\\
\hline
P@N ($\%$)& $10\%$ &$20\%$ &$50\%$&F1&$10\%$ &$20\%$ & $50\%$&F1&$10\%$ & $20\%$ &  $50\%$&F1\\
\hline
CNN+rand & 86.7&67.0& 38.9& 57.5& 84.6& 66.4& 37.5&55.0&79.9& 61.8&35.2 &51.4\\
CNN+max  & 86.0& 68.5& 38.3& 57.2 & 85.4& 67.6& 37.7&56.5&84.4&66.0 & 36.6&54.8\\
\textbf{Path+rand} & \textbf{89.4}& \textbf{71.7}& \textbf{39.9}& 59.3& 88.2& 70.2& 39.0&58.1&86.0& 67.2& 37.0&55.6\\
\textbf{Path+max} & 89.0& 71.5& 39.8& \textbf{59.6}&\textbf{89.0} & \textbf{71.4}& \textbf{39.6}&\textbf{59.4}&
\textbf{88.6} &\textbf{71.0} & \textbf{39.1}&\textbf{59.1}\\
\hline
\end{tabular}
\caption{\label{table:robust}P@N and F1 for relation extraction in texts containing different percentage of no-relation facts.}
\end{table*}

\subsubsection{Precision-Recall Curve Comparison}
To demonstrate the effect of our approach, we empirically compare it with other neural relation extraction methods via held-out evaluation. (1) \textbf{CNN+rand} represents the CNN model reported in \cite{zeng2014relation}. (2) \textbf{CNN+max} represents the CNN model with multi-instances learning used in \cite{zeng2015distant}. (3) \textbf{Path+rand/max} is our model with those two multi-instance settings. We implement (1), (2) by ourselves which achieve comparable results as reported in those papers. 

\begin{figure}[!htp]
\centering
\includegraphics[width=1.0\columnwidth]{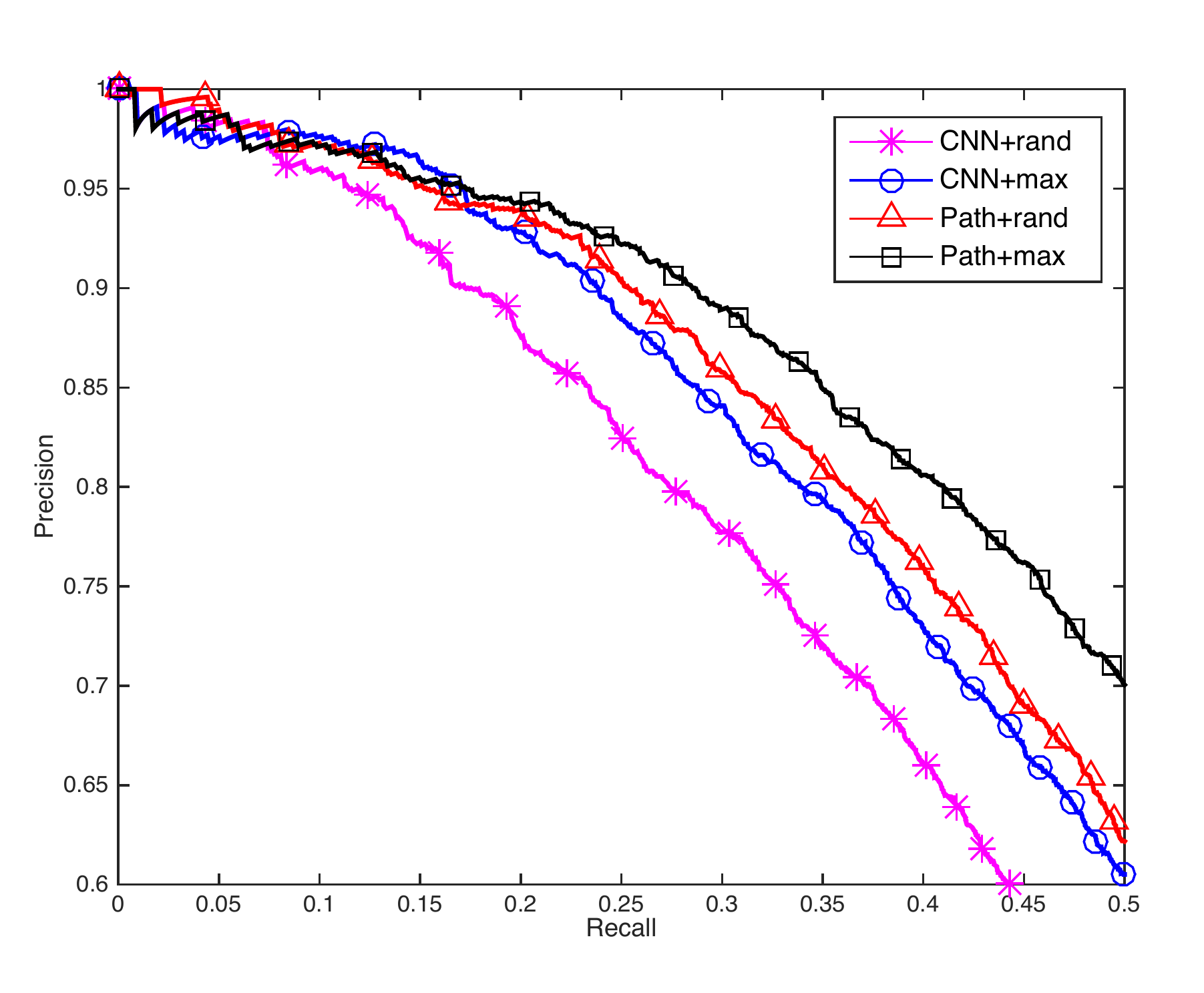}
\caption{Aggregate precision/recall curve for CNN+rand, CNN+max, Path+rand, Path+max.}\label{fig:1}
\label{fig:exp-main}
\end{figure}

Fig. \ref{fig:exp-main} shows the precision/recall curves of all methods. From the figure, we can  observe that: (1) Our methods outperform their counterpart methods, achieving higher precision over almost entire range of recall. They also enhance recall by 20$\%$ without decrease of precision. These results prove the effectiveness of our approach. We notice that the improvements of our methods over baselines are relatively small at small recall value, which corresponds to high predicting confidence. This phenomenon is intuitive since our joint model could dynamically leverage the importance of direct sentence and relation paths, and tends to trust the Text Encoder when the confidence is high. (2) As the recall increases, our models exhibit larger improvements compared with CNN in terms of percentage. This is due to the fact that sometimes CNNs  cannot extract reliable information from direct sentences, while our methods could alleviate this issue by considering more information from inference chains, and thus still maintain high precision. (3) Both CNN+max and Path+rand are variations of CNN+rand, aiming to alleviate the problem of noisy data. We see that Path+rand outperforms CNN+max over all range, which indicates that considering path information is a better way to solve this issue. Meanwhile, combining paths information and max operation, Path+max, gives the best performance. (4) Path+rand shows a larger improvement over CNN+rand, compared with those of Path+max and CNN+max. This furthermore proves the effectiveness of considering relation path information: CNN+rand has much more severe problem suffering from noise, so using our method to incorporate paths information to alleviate this issue could perform better.

\subsubsection{Comparison on Long Tail Situation}

\begin{table}[!htp]
\centering
\tiny
\begin{tabular}{c|cccc}
\hline
		&	$N_s \leq 1$	&$N_s \leq 2$	&$N_s \leq 5$	&All	\\
\hline
CNN+rand& 	53.9&	54.0&	52.0&	51.4\\
\textbf{Path+rand}& 	\textbf{58.4}&	\textbf{58.1}&	\textbf{56.0}&	\textbf{55.6}\\
\hline
CNN+max& 	57.8&	58.3&	56.5&	55.7\\
\textbf{Path+max}& 	\textbf{63.6 (+5.8)}&	\textbf{62.5 (+4.2)}&	\textbf{60.2 (+3.7)}&	\textbf{59.1 (+3.4)}\\
\hline
\end{tabular}
\caption{F1 score for long tail situation.}
\label{table:frequency}
\end{table}

Real-world data follows long-tail distribution (power law). In the testing set, we also observe a fact that about $40 \%$ triple facts appear only in single sentence, and thus a multi-instance relation extraction system, e.g. CNN+max, could only rely on limited information and the multi-instance mechanism will not work well. Our system, on the contrary, can still utilize information from relation paths in this case, and is expected to perform much better in the long tail situation. 

We evaluate the models on different parts of the long-tail distribution. To get testing instances from different parts of the distribution, we extract all the triple facts appearing less than $N_s$ sentences in the testing set, and those sentences associating with them. All text related to 'No Relation' entity pair are also reserved in order to simulate noise. We then evaluate different models on those sampled testing set, and report the results of F1 score in Table \ref{table:frequency}.

From Table \ref{table:frequency}, we could observe that: (1) Incorporating relation paths is indeed effective in predicting relations, and our models have significant improvements compared with the baselines. (2) Path+max indeed has larger improvements over CNN+max when $N_s$ is small, which is consistent with our previous expectation. Also notice that the gap between Path+rand and CNN+rand is relatively constant. This is due to the fact that both these methods only use one random sentence, regardless of how many sentences there are associating with an entity pair.

\begin{table*}[htp]
\centering
\small
\begin{tabular}{|c|c|c|}

\hline
&Relation & Text\\
\hline
Path $\#1$ & mother&\textbf{Rebecca} gave birth to twin sons, \textbf{Esau} and Jacob, ...\\
\hline
Path $\#2$ & has$\_$child	 & ...\textbf{Isaac}'s marriage to Rebecca, by whom he has two sons, \textbf{Esau} and jacob, ...\\
\hline
Test & spouse& ... \textbf{Isaac} and \textbf{Rebecca} and the female and male evil spirits ... \\
\hline
Path $\#1$ & shares$\_$border$\_$with& ... in \textbf{Somalia}, ... soldiers and marines stationed in neighboring \textbf{Djibouti} ...\\
\hline
Path $\#2$ & shares$\_$border$\_$with&... \textbf{Ethiopia} have had the effect of making neighboring \textbf{Djibouti} ...\\
\hline
Test & shares$\_$border$\_$with&The next day, \textbf{Ethiopia} struck, its military pushing deep into \textbf{Somalia} ...\\
\hline

\end{tabular}
\caption{Some representative examples of inference chians in NYT corpus. The bold is target entities. }
\label{table:case}
\end{table*}

\subsection{Model  Robustness under Different Percentages of Noise}

In the task of relation extraction, there are lots of noise in text which may hurt the model's performance. More specifically, ``No Relation" entity pair is a kind of noise, since ``No Relation" could actually contain many unknown relation types, and thus might confuse the relation extraction systems. 
Therefore, it is important to verify the robustness of our model in the presence of massive noise. 
Here, we evaluate those models in three settings, with the same relational facts and different percentages of ``No Relation" sentences in the testing sets. In each experiment, we extract top 20,000 predicting relational facts according to the model's predicting scores, and report the precision @top 10$\%$, @top 20$\%$, @top 50$\%$ and F1 score in Table \ref{table:robust}. 

From the table, we can see that:  (1) In terms of all evaluations, our models achieve the best performance as compared with other methods in all test settings. It demonstrates the effectiveness of our approach. (2) Even though the scores of all models drop as the increasing of noise, we find that Path+rand/max's scores decrease much less than their counterparts. This result proves the effectiveness of taking inference chains into consideration. Since we utilize more information to make predictions, our model is more robust to the presence of mass noise. 

\subsection{Effectiveness of Learned Features in Zero-Shot Scenario}

It has been proved that CNN could automatically extract useful features, encoding syntactic and semantic meaning of sentences. These features are sometimes fed to subsequent models to solve other tasks. In this experiment, we demonstrate the effectiveness of the extracted features from our model. Since CNN-based models have already succeeded in extracting relations from single sentences, we set our experiment in a new scenario: predicting the relation between entities which have not appeared in the same sentence. 

A natural approach is to build a relation path between this zero-shot entity pair. We assume that we can make a prediction about $(h,t)$, once we know the information of $(h,e)$ and $(e,t)$. Therefore, we build the training set by extracting all such relation paths and their sentences from training text, and similar for testing set. To test the effectiveness of features, we encode sentences by CNN+rand/max, Path+rand/max respectively, and then feed the concatenation of sentence vectors to a logistic classifier.

\begin{table}[!htp]
\centering
\small
\begin{tabular}{cc}
\hline
Feature & Accuracy\\
\hline
CNN+rand & 56.9\\
CNN+max & 57.3\\
\textbf{Path+rand} & 58.5\\
\textbf{Path+max} &  \textbf{60.4}\\
\hline

\end{tabular}
\caption{Accuracy of different models in zero-shot situation.}
\label{table:zero}
\end{table}

From Table \ref{table:zero}, we could observe that: (1) The result using CNN+rand features is comparable to the result using CNN+max features. It shows that using max operation to train the features does not greatly improve the features' behavior in this task, even though it performs well in previous tasks. The reason is that, both CNN+rand and CNN+max only encode the information from a single sentence, and they are unable to capture the correlations between relations. (2) Feature from Path+rand/max shows its effectiveness over those from other methods. It indicates that our method is able to model the correlations between relations, while also keeps the syntactic and semantic meaning of a sentence. Therefore, the features extracted from Path+rand/max are useful for a wider range of applications, especially in those tasks which need the information from relations.

\subsection{Case study}

Table \ref{table:case} shows some representative inference chains from the testing dataset. These examples can not be predicted correctly by the original CNN model, but are later corrected using our model. We show the test instances and their correct relations, as well as the inference chains the model uses. In the first example, the test sentence does not directly express the relation \texttt{spouse}, the proof of this relation appears in a further context in NYT. However, using path\#1 and path\#2, we could easily infer that \emph{Rebecca} and \emph{Issac} are \texttt{spouse}. The second example doesn't show the relation  either. But with the help of intermediate entity, \emph{Dijibouti}, our model predicts that \emph{Somalia} shares the border with \emph{Ethiopia}. Note that this inference chain doesn't always hold, but our model could capture this uncertainty well via a softmax operation. In general, our model can utilize common sense from inference chains. It helps make correct predictions even if the inference is not explicit.

\section{Conclusion and Future Work}

In this paper, we propose a neural relation extraction model which encodes the information of relation paths. As compared to existing neural relation extraction models, our model is able to utilize the sentences which contain both two target entities and only one target entity and is more robust for noisy data.  Experimental results on real-world datasets show that our model achieves significant and consistent improvements on relation extraction as compared with baselines.

In the future, we will explore the following directions: \textbf{(1)} 
We will explore the combination of relation paths from both plain texts  and KBs for relation extraction. \textbf{(2)}
We may take advantages of probabilistic graphical model or recurrent neural network to encode more complicated correlations between relation paths, e.g. multi-step relation paths, for relation extraction.

\section*{Acknowledgments}

This work is supported by the 973 Program (No. 2014CB340501), the National Natural Science Foundation of China (NSFC No. 61572273, 
61661146007), China Association for Science and Technology
(2016QNRC001), and Tsinghua University Initiative Scientific Research
Program (20151080406).

\bibliography{my}
\bibliographystyle{emnlp_natbib}

\end{document}